\pdfoutput=1
\documentclass[11pt]{article}

\usepackage[final]{acl}
\usepackage{times}
\usepackage{latexsym}
\usepackage[T1]{fontenc}
\usepackage[utf8]{inputenc}
\usepackage{microtype}
\usepackage{inconsolata}
\usepackage{graphicx}
\usepackage{amsfonts,amsmath,amssymb,amsthm}
\usepackage[ruled,vlined]{algorithm2e}
\usepackage{algorithmic}
\usepackage{booktabs}
\usepackage{float}
\usepackage{multicol,multirow}
\usepackage[most]{tcolorbox}
\usepackage{graphicx}
\usepackage{lipsum,afterpage,refcount}
\newcommand{\setfootnotemark}{%
  \refstepcounter{footnote}%
  \footnotemark[\value{footnote}]}

\global
\tcbset{
  aibox/.style={
    width=\textwidth,
    top=10pt,
    colback=white,
    colframe=black,
    colbacktitle=black,
    enhanced,
    center,
    attach boxed title to top left={yshift=-0.1in,xshift=0.15in},
    boxed title style={boxrule=0pt,colframe=white,},
  }
}
\newtcolorbox{AIbox}[2][]{aibox,title=#2,#1}
\tcbset{
  aiboxsmall/.style={
    width=0.96\textwidth,
    top=10pt,
    colback=white,
    colframe=black,
    colbacktitle=black,
    enhanced,
    center,
    attach boxed title to top left={yshift=-0.1in,xshift=0.15in},
    boxed title style={boxrule=0pt,colframe=white,},
  }
}
\newtcolorbox{AIboxSmall}[2][]{aiboxsmall,title=#2,#1}
\definecolor{aigold}{RGB}{244,210, 1} 
\definecolor{aired}{RGB}{255,180,181}
\newlength\savewidth

\definecolor{defaultcolor}{gray}{0.9}

\newcommand{\ours}{T-REG}

\title{
T-REG: Preference Optimization with Token-Level Reward Regularization
}

\author{Wenxuan Zhou, Shujian Zhang, Lingxiao Zhao, Tao Meng\\
Zoom Communications\\
\texttt{wenxuan.zhou@zoom.us}}

\begin{document}
\maketitle
\begin{abstract}
Reinforcement learning from human feedback (RLHF) has been crucial in aligning large language models (LLMs) with human values.
Traditionally, RLHF involves generating responses to a query and using a reward model to assign a reward to the entire response.
However, this approach faces challenges due to its reliance on a single, sparse reward, which makes it challenging for the model to identify which parts of the sequence contribute most significantly to the final reward.
Recent methods have attempted to address this limitation by introducing token-level rewards.
However, these methods often rely on either a trained credit assignment model or AI annotators, raising concerns about the quality and reliability of the rewards.
In this paper, we propose \underline{t}oken-level reward \underline{reg}ularization ({\ours}), a novel approach that leverages both sequence-level and token-level rewards for preference optimization.
Harnessing the self-refinement capabilities of LLMs, our method uses contrastive prompting to enable LLMs to self-generate token-level rewards.
These self-generated rewards then act as reward regularization, guiding the model to more effectively distribute sequence-level rewards across tokens.
This facilitates better token-level credit assignment and enhances alignment performance.
Experiments on the instruction following benchmarks, including Alpaca Eval 2 and Arena-Hard, show that our method consistently outperforms baseline methods by up to 3.8\% and 4.4\%, respectively.\footnote{We will release the code and models at \url{https://github.com/wzhouad/T-REG}.}

\end{abstract}

\section{Introduction}
% Main motivation:
% 1. Preference is usually given on sequence or step-level, hard to ensure correct credit assignment to individual tokens, while annotation on individual tokens is impossible.
% 2. DPO implicitly learns the token-level reward from sequence-level supervision.
% 3. Inspired by contrastive decoding, we utilize LLMs to automatically generate (weak) token-level supervision, and use it to guide the learning of DPO to learn better credit assignment.

% \sz{working}

% Reinforcement learning from human feedback \cite{}
Recent advancements in large language models (LLMs; \citealt{tunstall2023zephyr, chung2024scaling, team2024gemma}) have centered on aligning model outputs with human intentions and preferences. Reinforcement learning from human feedback (RLHF; \citealt{christiano2017deep, glaese2022improving}) has become a dominant approach for achieving this alignment by incorporating human feedback into the training process.
The RLHF process typically consists of two phases. In the first phase, responses are generated either with the model being optimized (on-policy RL; \citealt{schulman2017proximal, guo2024direct}) or with different models (off-policy RL; \citealt{rafailov2024direct, ethayarajh2024kto}).
In the second phase, a reward is obtained based on feedback from either humans~\cite{schulman2017proximal} or AI annotators~\cite{lee2023rlaif}, and this reward is then used to optimize the policy model through algorithms such as PPO~\cite{schulman2017proximal} and DPO~\cite{rafailov2024direct}.
During this process, the quality of the reward signal is critical to the success of RLHF, where poor or misaligned rewards can lead to issues such as reward hacking~\cite{gao2023scaling} and model collapse~\cite{wang2024secrets, chowdhury2024provably}.

\begin{figure*}[t]
\centering
\includegraphics[width=16cm]{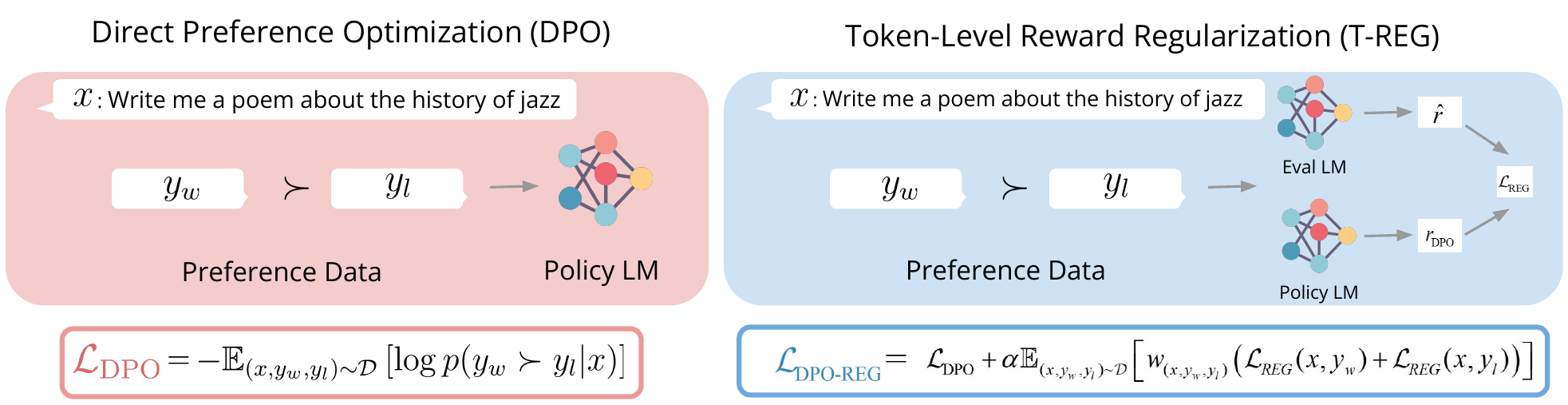}
\caption{Overview of the Token-Level Reward Regularization ({\ours}). Existing DPO directly optimizes the sequence-level rewards to align with user preferences. {\ours} prompts LLMs to generate the token-level reward and utilize it as regularization during the preference optimization. 
}
\label{fig:pipeline}
\end{figure*}

Current RLHF methods typically use sequence-level rewards, where reward signals are provided at the end of an entire sequence. Recent studies have shown that incorporating finer-grained reward signals~\cite{wu2024fine, lightman2023let}, ultimately at the token level~\cite{zhong2024dpo}, can significantly enhance the performance and convergence of the policy model.
However, annotating token-level rewards is challenging for humans, especially when sequences may span thousands of tokens.
As a result, current token-level RLHF methods often rely on token-level rewards labeled by AI annotators~\cite{guo2023beyond, yoon2024tlcr} or utilize credit assignment models that learn to redistribute sequence-level rewards to individual tokens~\cite{yang2024preference, zhong2024dpo}.
These token-level rewards are then used to train policy models through algorithms such as PPO.
However, the quality of these automatically generated rewards remains a concern. AI-generated annotations and credit assignment models are prone to errors, raising questions about the reliability of the rewards and, consequently, the overall quality of the trained policy models.

In this paper, we investigate the integration of auto-labeled token-level rewards into preference optimization.
We introduce {\ours}, a novel token-level preference optimization algorithm that leverages both sequence-level and token-level rewards. Our approach is motivated by the observation that DPO inherently learns token-level rewards.
Instead of directly optimizing the policy model with auto-labeled token-level rewards, {\ours} retains sequence-level preference optimization while using these token-level rewards as weak supervision.
This weak supervision regularizes the token-level reward signals implicitly learned during preference optimization.
This design enables {\ours} to optimize the overall sequence-level reward while also achieving effective token-level credit assignment.
Inspired by the self-refinement capabilities of LLMs~\cite{madaan2024self}, we derive token-level rewards through contrastive prompting, where revision-based prompts rewrite the output in opposing directions.
By comparing the token probabilities generated from these prompts, we compute the token-level rewards. This approach assigns token-level rewards effectively without requiring additional training or token-level annotations.

We evaluate {\ours} on two instruction following benchmarks, including Alpaca Eval 2~\cite{alpaca_eval} and Arena-Hard~\cite{li2024crowdsourced}.
{\ours} demonstrates consistent improvements across both Llama-3-Instruct-8B and Gemma-2-9B-it models.
On Alpaca Eval 2, {\ours} increases the length-controlled win rate by up to 24.8\% over the SFT checkpoint, surpassing DPO by up to 3.8\%. On the more challenging Arena-Hard benchmark, {\ours} improves the win rate by up to 20.0\%, outperforming DPO by as much as 4.4\%.
Moreover, we find that {\ours} can be integrated into different preference optimizations (e.g., SimPO) and shows consistent improvements.
Additionally, we analyze the token-level rewards learned by {\ours}, demonstrating that it effectively achieves token-level credit assignment.

Our contributions are summarized as follows:
\begin{itemize}
\setlength{\itemsep}{0pt}
\setlength{\parsep}{0pt} 
\setlength{\parskip}{0pt}
\item We propose {\ours}, a token-level preference optimization method that utilizes both sequence-level and token-level rewards to enhance alignment with human preferences.
\item {\ours} utilizes token-level rewards derived through contrastive prompting to guide the token-level rewards learned during preference optimization, enabling effective token-level credit assignment without the need for external token-level reward annotations.
\item {\ours} achieves consistent improvements on instruction-following benchmarks including Alpaca Eval 2 and Arena-Hard, outperforming DPO by up to 3.8\% and 4.4\%, respectively, and showing compatibility with other preference optimization algorithms.
\end{itemize}

\section{Related Work}

\paragraph{Reinforcement learning from human
feedback.}
Reinforcement Learning from Human Feedback (RLHF) has become a cornerstone in aligning Large Language Models (LLMs) with human intentions, allowing for fine-tuning of models to produce more useful and aligned responses \cite{ziegler2019fine}. The early implementations of RLHF typically used the proximal policy optimization algorithm (PPO; \citealt{schulman2017proximal}), which operates by optimizing policies through episodic rewards, calculated using a reward model trained on human preferences \cite{bai2022training, ouyang2022training}. However, applying deep RL techniques such as PPO to LLMs has proven challenging due to instability and inefficiency in training \cite{engstrom2019implementation}. To address these challenges, alternative methods have emerged that employ supervised fine-tuning (SFT) to simulate the RL process, using high-quality responses to create more stable training signals \cite{lu2022quark, zhao2023slic, yang2024preference}. 
In contrast to classical RLHF methods, a new line of research has proposed direct alignment algorithms~\cite{zhao2023slic,rafailov2024direct,meng2024simpo,azar2024general,zhou2024wpo,wang2024mdpo}, which optimize a supervised target directly based on preference data, without the need for a separate reward model.
Despite their advantages, these methods rely solely on sequence-level rewards.
In this work, we extend this paradigm by prompting LLMs to generate token-level rewards using contrastive prompting, integrating these rewards as regularization during preference optimization.

\smallskip
\noindent\textbf{Token-level RLHF.}
While sequence-level rewards provide effective supervision for RLHF, the extended length of LLM generations means that different parts of a sequence contribute unequally to the final rewards, making sequence-level rewards less data-efficient and effective \cite{chan2024dense, zhong2024dpo}.
To address these limitations, recent efforts have shifted from sequence-level to token-level rewards, with a focus on how to effectively derive these finer-grained signals.
Existing methods have adopted various approaches to token-level reward derivation.
For instance, Token-Level Direct Preference Optimization (TDPO; \citealt{zeng2024token}) uses Markov decision processes (MDPs) to derive token-level rewards.
Other methods, such as token-level continuous reward (TLCR; \citealt{yoon2024tlcr}), FIGA \cite{guo2023beyond}, and RLMEC \cite{chen2024improving}, rely on off-the-shelf LLMs to auto-label token-level rewards.
Additionally, since DPO has been shown to implicitly learn token-level rewards~\cite{rafailov2024direct}, some approaches~\cite{zhong2024dpo, yang2024selective} leverage DPO for this purpose.
Once token-level rewards are derived, existing methods typically optimize the policy using algorithms such as PPO \cite{zhong2024dpo, yoon2024tlcr} or token selection strategies \cite{yang2024selective}.
However, these approaches often depend heavily on auto-labeled token-level rewards, which introduces potential challenges due to the inherent noise of auto-labeled token-level rewards.
In our approach, we use token-level rewards derived from LLMs as guidance in preference optimization to enhance token-level credit assignment.
Therefore, we mitigate the risks of solely relying on auto-labeled token-level rewards, achieving a more robust and effective optimization process.

\section{Method}
In this section,  we present in detail our proposed
method, {\ours} (Algorithm \ref{alg:t_reg}). We first provide the theoretical background of RLHF in Section~\ref{ssec:preliminaries}. Next, we introduce the token-level credit assignment problem and explain how to guide preference optimization using a token-level regularization in Section~\ref{ssec:main_method}. Finally, we present our contrastive prompting method for generating token-level rewards in Section~\ref{ssec:self_token_reward}.

\begin{algorithm}[t]
\caption{Token-Level Reward Regularization (T-REG)} 
\label{alg:t_reg}
\begin{algorithmic}
\STATE \textbf{Input:} Dataset ($\mathcal{D}$) with prompts and respon-\\ses, policy model $\pi_{\theta}$, total number of iterations \\$T$, learning rate $\alpha_t$,
\FOR{$t =0$  to $T$}
\STATE Sample a mini-batch of tuples $(x, y_w, y_l)$ from $\mathcal{D}$,
\STATE Generate token-level rewards via Eq. 
\eqref{eq:self_generated_reward},
\STATE Define the regularization at token-level rewards via Eq. 
\eqref{eq:regularization_loss},
\STATE Compute  $\mathcal{L}_\text{DPO-REG}$ via Eq. \eqref{eq:dpo_reg_loss}, 
\STATE Update policy parameters $\theta$ using gradient descent: $\theta  \leftarrow \theta  -  \alpha_t\nabla \theta(x, y_w, y_l, \theta)$. 
 \ENDFOR
\end{algorithmic}
\vspace{-2pt}
\end{algorithm}

\subsection{Preliminaries}
\label{ssec:preliminaries}
\smallskip
\noindent\textbf{RLHF.}
Reinforcement learning from human feedback (RLHF; \citealt{schulman2017proximal}) is a widely used technique for aligning LLMs with human preferences. Using human or AI annotations, RLHF enables LLMs to generate outputs that align better with human expectations.
Consider a preference dataset $\mathcal{D} = {\left(x^{(i)}, y_w^{(i)}, y_l^{(i)}\right)}_{i=1}^N$, where each sample consists of a prompt $x^{(i)}$ and two outputs, $y_w^{(i)}$ (preferred) and $y_l^{(i)}$ (dispreferred), as judged by annotators, the core of RLHF lies in a reward model, $r^*(x, y)$, which assigns a score to each candidate output $y$ based on how well it aligns with the prompt $x$.
A commonly adopted framework for modeling preference distributions is the Bradley-Terry (BT; \citealt{bradley1952rank}) model, which defines the probability that $y_w$ is preferred over $y_l$ as:
\begin{equation*}
\resizebox{0.46\textwidth}{!}{ $p(y_w \succ y_l|x)=\frac{\exp(r^*(x, y_w))}{\exp(r^*(x, y_w)) + \exp(r^*(x, y_l))}$}.
\end{equation*}
The RLHF process typically consists of two phases: learning the reward model from preference data, followed by using reinforcement learning to optimize a policy model based on this reward.
In the first phase, the reward model is trained using maximum likelihood estimation, producing an estimated reward function $\hat{r}(x, y)$.
The reward model can be structured to return feedback either at the end of a sequence~\cite{liu-etal-2023-g,dong2023raft,xiong2024iterative}, where the evaluation is based on the entire output, or at each step of the sequence~\cite{uesato2022solving,lightman2023let}, where feedback is provided based on intermediate reasoning steps.
Once the reward model is trained, it is used to finetune the policy model by optimizing the following objective:
\begin{equation*}
\resizebox{0.46\textwidth}{!}{
    $\max_{\pi_\theta} \mathbb{E}_{x\sim \mathcal{D},y\sim \pi_\theta(\cdot|x)}\left[\hat{r}(x,y)-\beta\log\frac{\pi_\theta(\cdot|x)}{\pi_\text{ref}(\cdot|x)}\right]$,}
\end{equation*}
where $\beta$ is a KL penalty coefficient to regularize the deviation between the policy model $\pi_\theta$ and the reference model $\pi_\text{ref}$.

\smallskip
\noindent\textbf{DPO implicitly learns token-level rewards.}
In RLHF, rewards are typically assigned at the end of sequences, which can extend to thousands of tokens.
However, not all tokens contribute equally to the final reward \cite{chan2024dense, rafailov2024r, yang2024selective}.
To address this, it is crucial for a preference optimization algorithm to effectively distribute sequence-level rewards across individual tokens.
Previous work \cite{zhong2024dpo, rafailov2024r} has shown that Direct Preference Optimization (DPO; \citealt{rafailov2024direct}) implicitly learns token-level rewards under sequence-level reward supervision.
Specifically, with the token-level reward defined as $r^*_\text{token}(y_t|x,y_{<t})=\beta\log\frac{\pi^*(y_t|x,y_{<t})}{\pi_\text{ref}(y_t|x,y_{<t})}$, the sequence-level reward is computed as\footnote{For a detailed derivation, refer to~\citet{rafailov2024r}.}:
\begin{equation*}
r_\text{DPO}(x,y)=V^*(x)+\sum_{t=1}^{T}\beta\log\frac{\pi^*(y_t|x,y_{<t})}{\pi_\text{ref}(y_t|x,y_{<t})},
\end{equation*}
where $T$ is the number of tokens in the sequence, $\pi^*$ is the optimal policy, $\pi_\text{ref}$ is the reference policy, and $V^*$ is the value function of $\pi^*$.
For a pair of outputs $(y_w, y_l)$, the probability that $y_w$ is preferred over $y_l$, modeled using the BT model, is given by:
\begin{align*}
p(y_w \succ y_l|x)=&\sigma\left(\sum_{t=1}^{T_w}\beta\log\frac{\pi^*(y_{w_t}|x,y_{w_{<t}})}{\pi_\text{ref}(y_{w_t}|x,y_{w_{<t}})}\right.\\
&\left.-\sum_{t=1}^{T_l}\beta\log\frac{\pi^*(y_{l_t}|x,y_{l_{<t}})}{\pi_\text{ref}(y_{l_t}|x,y_{l_{<t}})}\right),
\end{align*}
where $V^*(x)$ cancels out because $y_w$ and $y_l$ correspond to the same prompt $x$.
By applying maximum likelihood estimation to the preference dataset, the policy model $\pi$ can be optimized by the following loss function:
\begin{equation*}
    \mathcal{L}_\text{DPO}=-\mathbb{E}_{(x, y_w, y_l)\sim \mathcal{D}} \left[ \log p(y_w \succ y_l|x)\right],
\end{equation*}
which resembles the original DPO loss~\cite{rafailov2024direct}.
Therefore, performing RLHF with DPO is redistributing the sequence-level reward into the token level as $\beta\log\frac{\pi(y_t|x,y_{<t})}{\pi_\text{ref}(y_t|x,y_{<t})}$, where $\pi$ is the policy trained by DPO. 
This redistribution ensures token-level credit assignment while optimizing the sequence-level objective.

\begin{figure*}[!ht]
    \begin{AIboxSmall}{\footnotesize Prompting Template for Revising an Output}
    \footnotesize
    \textbf{Instruction:} Below is a conversation between an user and an AI Assistant.

[User Question]

\{instruction\}

[The start of Assistant’s Answer]

\{answer\}

[The end of Assistant’s Answer]

Please rewrite the Assistant’s Answer to make it \{direction\}. Specifically, the rewritten \{direction\} answer should closely resemble the original answer but is \{direction\} in terms of one or multiple of the following aspects:

helpfulness, correctness, coherence, verbosity.

IMPORTANT: Please strictly follow the following format:
First, choose one or multiple aspects to generate a \{direction\} answer, such as rewrite the original answer to be \{detailed\_description\}, etc.

[The start of a rewritten \{direction\} answer]

<provide a \{direction\} answer here>

[The end of a rewritten \{direction\} answer]
    \end{AIboxSmall}
    \vspace{-10pt}
    \caption{Prompt template for revising the output to either a better or worse one. To make the output \textit{better}, we set use ``helpful, correct, coherent, concise'' as the description. To make the output \textit{worse}, we use ``unhelpful, incorrect, incoherent, verbose'' as the description.}
    \label{fig:prompt_revision}
\end{figure*}

\subsection{Regularized Token-level Preference Optimization}
\label{ssec:main_method}

\smallskip
\noindent\textbf{Token-level reward as regularization.}
In preference optimization, our objective is to ensure consistency between the model's pairwise rankings and human preference while also achieving effective token-level credit assignment to enhance the model's generalization capabilities.
Since DPO models are trained with sequence-level rewards as supervision, they can capture pairwise rankings at the sequence level~\cite{lambert2024rewardbench}.
However, since token-level rewards are implicitly derived through the redistribution of sequence-level rewards in DPO, they often lack direct guidance.
Recent studies have demonstrated that LLMs can serve as dense token-level reward functions without fine-tuning, by employing techniques such as contrastive decoding~\cite{li2022contrastive} or contrastive prompting~\cite{kim2024instructive,zhao2024adversarial}.
These methods infer token-level rewards from the difference in token probabilities between a strong model and a weak model, expressed as $\log\frac{\pi_\text{strong}(y_t|x,y_{<t})}{\pi_\text{weak}(y_t|x,y_{<t})}$ (see details in Section~\ref{ssec:self_token_reward}).
However, these methods do not guarantee that the accumulated sequence-level rewards align with the ground-truth sequence-level rewards provided by the preference data.
In this paper, we propose integrating token-level and sequence-level rewards to leverage the strengths of both. Our approach incorporates token-level rewards as regularization in preference optimization to improve token-level credit assignment, thereby improving generalization of the policy model.

Specifically, we introduce a regularization term to ensure that the token-level rewards learned by DPO align with dense token-level rewards derived from LLMs.
Let $r_\text{token}$ denote the token-level reward learned by policy $\pi$, and $\hat{r}_\text{token}$ denote the given dense token-level reward, we can compute their similarity at $y_t$ by:
\begin{equation*}
\resizebox{0.46\textwidth}{!}{$
    \text{sim}(y_t\ |\ x,y_{<t}) = r_\text{token}(y_t\ |\ x,y_{<t}) \, \hat{r}_\text{token}(y_t\ |\ x,y_{<t}).$}
\end{equation*}
Our goal is to maximize the alignment between the token-level rewards across the entire output.
The regularization term, therefore, is formulated as:
\begin{align*}
    \mathcal{L}_\text{reg}&= -\sum_{t=1}^{T}\text{sim}(y_t\ |\ x,y_{<t}) \\
    &\resizebox{0.44\textwidth}{!}{$=-\sum_{t=1}^{T}\beta\hat{r}_\text{token}(y_t\ |\ x,y_{<t})\,\log\frac{\pi(y_t|x,y_{<t})}{\pi_\text{ref}(y_t|x,y_{<t})}.$}
\end{align*}
Here, the term $\pi_\text{ref}(y_t|x,y_{<t})$ is a constant reference probability, which does not affect the gradient and can thus be omitted.
This leads to the following modified regularization term:
\begin{equation}
\label{eq:regularization_loss}
\resizebox{0.42\textwidth}{!}{$
    \mathcal{L}_\text{reg}=-\sum_{t=1}^{T}\beta\hat{r}_\text{token}(y_t\ |\ x,y_{<t})\,\log \pi(y_t|x,y_{<t}).$}
\end{equation}
This term acts as a weighted language modeling loss, increasing the likelihood of tokens with positive token-level rewards while decreasing the likelihood of tokens with negative token-level rewards, thereby improving token-level credit assignment.

\smallskip
\noindent\textbf{Regularized token-level preference optimization.}
During preference optimization, we then optimize both the preference optimization loss (e.g., $\mathcal{L}_\text{DPO}$) and the regularization loss ($\mathcal{L}_\text{reg}$).
The preference optimization loss aims to increase the margin between the probabilities of preferred and dispreferred sequences, while the regularization loss maximizes the probabilities of high-reward tokens.
Since these two objectives optimize in different directions, balancing them is critical for effective optimization.
To achieve this balance, we use sequence-level gradient norms~\cite{chen2018gradnorm}.
Recall that the gradient of the DPO loss is:
\begin{align*}
    \nabla \mathcal{L}_\text{DPO}&=-\beta \sigma\left(r_\text{DPO}(x, y_l) - r_\text{DPO}(x, y_w)\right) \\
    &\left(\nabla \log \pi\left(y_w|x\right)-\nabla \log \pi\left(y_l|x\right)\right).
\end{align*}
where for each sequence, the gradient norm with respect to each token is proportional to $\sigma\left(r_\text{DPO}(x, y_l) - r_\text{DPO}(x, y_w)\right)$.
To ensure that the regularization loss does not dominate the preference optimization loss for each sequence, we introduce a sequence weight to modulate the regularization loss.
The final regularized token-level preference optimization objective is:
\begin{align}
\begin{split}
    \mathcal{L}_\text{DPO-REG}=\mathcal{L}_\text{DPO}+\alpha\mathbb{E}_{(x, y_w, y_l)\sim \mathcal{D}}\left[w_{(x, y_w, y_l)}\right. \\
    \left.\left(\mathcal{L}_\text{REG}(x, y_w)+\mathcal{L}_\text{REG}(x, y_l)\right)\right],
\label{eq:dpo_reg_loss}
\end{split}
\end{align}
where $w_{(x, y_w, y_l)}=\sigma\left(r_\text{DPO}(x, y_l) - r_\text{DPO}(x, y_w)\right)$ is the sequence weight and is \textit{detached} from back propagation, $\alpha$ is a hyperparameter that controls the strength of regularization.
As shown in our ablation study (see Section~\ref{ssec:main_results}), sequence weighting achieves the best performance.

\subsection{Self-generated Token-level Rewards}
\label{ssec:self_token_reward}
In this section, we focus on deriving the dense token-level rewards by LLMs.
Existing work~\cite{zhou2024weak,zhao2024weak} often approximates this reward by contrasting a strong model against a weak model, where:
\begin{equation*}
\hat{r}(y_t|x,y_{<t})=\log\frac{\pi_\text{strong}(y_t|x,y_{<t})}{\pi_\text{weak}(y_t|x,y_{<t})}.
\end{equation*}
The strong model can be a model from the same family but at a larger scale~\cite{li2022contrastive}, a model that is more aligned with human expectations~\cite{zhou2024weak,zhong2024dpo,huang2024offset}, or a model with better prompts~\cite{kim2024instructive,zhao2024adversarial}.
In this paper, we focus on using \textit{contrastive prompting} to derive token-level reward, as this approach only requires a single model and does not require additional finetuning.
Specifically, we utilize two contrastive, revision-based prompts, $x_\text{better}$ and $x_\text{worse}$ designed to refine the current output $y$ in positive or negative directions to evaluate token quality.
Building on prior work in self-refinement~\cite{madaan2024self}, we leverage the demonstrated ability of LLMs to adjust outputs in diverse directions.
The revision prompt, adapted from~\citet{wang2024self} and illustrated in Fig.~\ref{fig:prompt_revision}, refines an output based on four aspects: helpfulness, correctness, coherence, and verbosity.
Given a token $y_t$ and a causal language model $\pi_\text{eval}$ for generating the rewards, we calculate the token-level reward by:
\begin{equation}
\label{eq:self_generated_reward}
\resizebox{0.42\textwidth}{!}{$
\hat{r}(x,y_{<t}, y_t) = \sigma\left(\log \frac{\pi_\text{eval}(y_t|x_\text{better},y_{<t})}{\pi_\text{eval}(y_t|x_\text{worse},y_{<t})}\right) - 0.5,$}
\end{equation}
where $\sigma$ clips the reward value, and $-0.5$ clips recenters the original reward into range of $[-0.5, 0.5]$.
This normalization helps mitigate extreme token-level reward values, thereby stabilizing the preference optimization process.
Thanks to the autoregressive nature of causal language models, the rewards for all tokens in an output can be derived with only two forward passes.
Due to tokenization issues, $\pi_\text{eval}$ should ideally share the same vocabulary as the reference model $\pi_\text{ref}$\footnote{Although there are methods to resolve mismatches in vocabulary~\cite{kasai2022twist}, we consider models with the same vocabulary for simplicity.}.
In this work, we specifically focus on the self-generated token-level rewards, where we use the reference model $\pi_\text{ref}$ to generate the token-level rewards.

\section{Experiments}
In this section, we outline our experimental settings in Section~\ref{ssec:experiment_settings}, present the main results and ablations in Section~\ref{ssec:main_results}, and provide qualitative case studies in Section~\ref{ssec:case_study}.

\subsection{Experimental Settings}
\label{ssec:experiment_settings}
\noindent\textbf{Model configurations.} Our methods are implemented using the official repo of Zephyr\footnote{\url{https://github.com/huggingface/alignment-handbook}}. Preference optimization is performed based on two large language models: Llama-3-8B-Instruct~\cite{dubey2024llama} and Gemma-2-9B-it~\cite{team2024gemma}.
We use the hyperparameters used by~\citet{meng2024simpo}, who conducted an extensive hyperparameter search. For our newly introduced hyperparameter, $\alpha$, we search in the range $\{0.1, 0.25, 0.5\}$.

\begin{table*}[!ht]
    \centering
    {
    \scalebox{0.77}{
    \begin{tabular}{lcccccc}
    \toprule
     \multirow{4}{*}{\textbf{Method}}& \multicolumn{3}{c}{\textbf{Llama-3-Instruct (8B)}}& \multicolumn{3}{c}{\textbf{Gemma-2-Instruct (9B)}}\\
     \cmidrule(lr){2-4}\cmidrule{5-7}
     &\multicolumn{2}{c}{\textbf{Alpaca Eval 2.0}}& \textbf{Arena-Hard}& \multicolumn{2}{c}{\textbf{Alpaca Eval 2.0}}& \textbf{Arena-Hard} \\
     \cmidrule(lr){2-3}\cmidrule(lr){4-4}\cmidrule(lr){5-6}\cmidrule(lr){7-7}
     &Len-control.& Win Rate& Win Rate& Len-control.& Win Rate& Win Rate\\
     &Win Rate& vs GPT-4& vs GPT-4& Win Rate & vs GPT-4& vs GPT-4 \\
     \midrule
     SFT& 26.0&25.3&22.3&51.1& 38.1&40.8\\
     \midrule
     RTO& 49.2& 47.2& 37.6& 67.6& 68.7& 63.1\\
     SePO& 48.5& 45.5& 33.1& -& -& -\setfootnotemark\label{first} \\
     TDPO$_1$& 45.4& 37.4& 29.1& 66.5& 56.6& 46.7\\
     TDPO$_2$& 43.2& 40.2& 34.0& 64.3& 59.3& 55.7\\
     \midrule
     DPO& 47.0& 46.0& 35.9& 68.9& 66.9& 58.6\\
     SimPO& 52.5& 47.1& 33.1& 73.5& \textbf{70.7}& 63.0\\
     \midrule
     DPO-REG& \underline{50.8}& \underline{\textbf{51.1}}& \underline{\textbf{40.3}}& \underline{70.3}& 66.4& 60.2\\
     SimPO-REG& \underline{\textbf{53.8}}& \underline{48.8}& 34.4& \underline{\textbf{74.5}}& 70.5& \underline{\textbf{64.2}}\\
     \bottomrule
    \end{tabular}}}
      \afterpage{\footnotetext[\getrefnumber{first}]{Despite extensive hyperparameter tuning, we find that Gemma-2-Instruct, fine-tuned by SePO, suffers severely from degeneration and fails to produce reasonable results on both benchmarks.}}
    \caption{Evaluation results (\%) on Alpaca Eval 2 and Arena-Hard benchmarks. Scores that are \underline{underlined} denote statistically significant gains ($p < 0.05$).}
    \label{tab:main_result}
    \vspace{-15pt}
\end{table*}

\smallskip
\noindent\textbf{Training data.}
We perform RLHF in an on-policy setting, where outputs are sampled from the policy being optimized.
However, generating outputs during training is computationally expensive.
To address this, we adopt an approximate method similar to that used by~\citet{meng2024simpo}, where outputs are sampled using the reference policy before preference optimization.
Specifically, the reference policy generates five outputs for each prompt, which are then evaluated using an external reward model.
The best and worst outputs are selected to form a preference pair.
For our experiments, we use the preference data generated by~\citet{meng2024simpo}, derived from prompts in Ultrafeedback~\cite{cui2023ultrafeedback} and scored using the ArmoRM reward model~\cite{ArmoRM, wang2024arithmetic}.

\smallskip
\noindent \textbf{Evaluation.}
We evaluate the models on two benchmarks: Alpaca Eval 2~\cite{alpaca_eval} and Arena-Hard~\cite{li2024crowdsourced}. Alpaca Eval 2 is an automated benchmark designed to assess the alignment of LLMs with human preferences across 805 representative instructions.
For each instruction, the evaluated model's response is compared head-to-head with the response generated by \texttt{gpt-4-turbo} using an automatic evaluator (with \texttt{gpt-4-turbo} as the default evaluator).
The win rate reflects the probability that the evaluator prefers the responses of the evaluated model over those of \texttt{gpt-4-turbo}.
Additionally, Alpaca Eval 2 introduces a length-controlled win rate~\cite{dubois2024length} to mitigate length bias in \texttt{gpt-4-turbo}.
We use the generation configurations recommended by~\citet{zheng2024weak} to generate the outputs during evaluation.

Arena-Hard is an automated benchmark consisting of 500 user queries that emphasize challenging topics, such as coding and expert-level knowledge.
The model's response is compared head-to-head with the response generated by \texttt{gpt-4-0314}.
Notably, Arena-Hard exhibits the highest correlation with Chatbot Arena and offers the best separability among widely used open-ended LLM benchmarks.
In our experiments, we use the official generation configurations provided by Arena-Hard.

\smallskip
\noindent\textbf{Compared methods.}
We consider two preference optimization algorithms, DPO and SimPO~\cite{meng2024simpo}\footnote{While T-REG is derived based on DPO, it can be adapted to other preference optimization algorithms by adjusting the sequence-level weighting term according to their gradients.}, as the base algorithms for applying {\ours}. Additionally, we evaluate the following token-level preference optimization algorithms for comparison:
(1) \textbf{RTO.} Reinforced token optimization~\cite{zhong2024dpo} first trains a DPO model to derive the dense token-level rewards. It then uses PPO~\cite{schulman2017proximal} to directly optimize the policy model based on these token-level rewards.
(2) \textbf{SePO.} Selective preference optimization~\cite{yang2024selective} also derives token-level rewards using DPO. It then performs preference optimization on a subset of tokens by selecting those with the highest reward values in the preferred output and the lowest reward values in the dispreferred output.
(3) \textbf{TDPO.} Token-level direct preference optimization~\cite{zeng2024token} optimizes preferences at the token level and includes forward KL divergence constraints for each token. This approach has two variants, TDPO$_1$ and TDPO$_2$. This method does not utilize explicit token-level rewards as supervision.
As these methods do not report their numbers on our base models, we reproduce them and search for the best set of hyperparameters within their recommended range.

\subsection{Main Results and Ablation}
\label{ssec:main_results}
\smallskip
\noindent\textbf{T-REG consistently outperforms the baselines and the compared methods.}
The results on Alpaca Eval 2 and Arena-Hard are presented in Table~\ref{tab:main_result}. We observe that {\ours}, when applied to both DPO and SimPO, consistently outperforms these methods on both benchmarks. Specifically, on Alpaca Eval 2, {\ours} increases the length-controlled win rate by up to 24.8\% over the SFT checkpoint, surpassing DPO by up to 3.8\%. On the more challenging Arena-Hard benchmark, {\ours} improves the win rate by up to 20.0\%, outperforming DPO by as much as 4.4\%.
Similar improvements are observed with SimPO, indicating that although {\ours} is primarily derived from DPO, it generalizes well to other preference optimization methods. Among other token-level preference optimization methods, RTO, which conducts PPO on token-level rewards derived from DPO, is the best performing one and achieves results comparable to or better than DPO, especially on the challenging Arena-Hard benchmark.
However, RTO's performance gains are smaller or even negative on Alpaca Eval 2, which focuses on general questions, while {\ours} consistently yields positive improvements across both benchmarks.
These results highlight the effectiveness of {\ours} in enhancing preference optimization.

\begin{table}[]
    \centering
    \scalebox{0.78}{
    \begin{tabular}{lccc}
    \toprule
        \multirow{3}{*}{\textbf{Method}}& \multicolumn{2}{c}{\textbf{Alpaca Eval 2.0}} & \textbf{Arena-Hard}\\
        \cmidrule(lr){2-3}\cmidrule(lr){4-4}
        &Len-control.& Win Rate& Win Rate\\
        &Win Rate& vs GPT-4& vs GPT-4 \\
        \midrule
        DPO & 47.0& 46.0& 35.9 \\
        DPO-REG& 50.8& 51.1& 40.3 \\
        \midrule
        DPO-SFT on $y_w$& 46.0& 43.4& 32.7 \\
        Static weigh.& 48.0& 46.3& 35.1 \\
        DPO reward& 49.8& 51.0& 36.9\\
        \bottomrule
    \end{tabular}}
    \caption{Results of different variants of {\ours} based on Llama-3-Instruct (8B).}
    \label{tab:ablation}
\end{table}

\smallskip
\noindent\textbf{Selective regularization on high-reward tokens yields better results.}
As shown in Eq.~\ref{eq:regularization_loss}, {\ours} can be interpreted as performing weighted SFT selectively focusing on high-reward tokens.
Previous methods~\cite{dubey2024llama,xucontrastive} apply an SFT loss over the entire preferred output $y_w$, which has been shown to help prevent degeneration:
\begin{equation*}
    \mathcal{L}_\text{DPO-SFT} = \mathcal{L}_\text{DPO} + \alpha \mathbb{E}_{(x, y_w, y_l) \sim \mathcal{D}} \left[\log\pi(y_w|x)\right].
\end{equation*}
However, as shown in Table~\ref{tab:ablation}, incorporating this loss results in significant performance degradation on both tasks, even with a small \( \alpha \) value.
This is likely because optimizing over the entire \( y_w \) includes tokens of lower quality.
In contrast, {\ours} — focuses exclusively on high-reward tokens — achieves consistent improvements over DPO.

\smallskip
\noindent\textbf{Sequence weighting enhances performance.}
To evaluate the impact of sequence-level weighting on performance, we trained DPO-REG using static weighting by removing the weighting term from the loss. However, as shown in Table~\ref{tab:ablation}, this approach did not result in consistent improvements over DPO, showing the effectiveness of sequence-level weighting.

\smallskip
\noindent\textbf{Self-generated reward outperforms DPO reward.}
In line with prior approaches like RTO and SePO, we also experimented with using the token-level rewards derived from DPO for regularization.
As shown in Table~\ref{tab:ablation}, this approach achieved similar performance to self-generated rewards on Alpaca Eval 2 but performed much worse on Arena-Hard, underperforming self-generated rewards by 3.4\%. These results indicate that self-generated token-level rewards are comparable to or better than those derived from DPO.

\smallskip
\noindent\textbf{Better preference data yields stronger results.}
To enhance our model further, we leverage hybrid preference data as proposed by~\citet{zhou2024wpo}.
This dataset is constructed by sampling five outputs from Gemma-2-9b-it and one additional output from \texttt{gpt-4-turbo}, followed by applying ArmoRM~\cite{wang2024arithmetic,ArmoRM} to identify the best and worst outputs, which are then used to form preference pairs.
Given the off-policy nature of this data, we utilize WPO, a method specifically designed for such scenarios, and apply {\ours} on top of WPO.
The model achieves a length-controlled win rate of 78.0\% on Alpaca Eval 2, compared to 76.7\% of WPO, demonstrating the effectiveness of the introduced approach.

\subsection{Case Study}
\label{ssec:case_study}
In this section, we examine the quality of token-level rewards learned by {\ours}. Since no existing evaluation datasets specifically assess token-level rewards, we perform a qualitative analysis instead. Figure~\ref{fig:case_study} presents three example tokens from Chatbot Arena~\cite{chiang2024chatbot}. We calculate the token-level rewards with \(\log \frac{\pi_\theta(y_t|x,y_{<t})}{\pi_\text{ref}(y_t|x,y_{<t})}\), where $\pi_\theta$ and $\pi_\text{ref}$ are the policy and reference models, respectively.
The positive and negative rewards are visualized in red and blue, respectively.
Our analysis demonstrates that integrating {\ours} into DPO enhances the precision of token-level reward assignments. 
\begin{itemize}
    \item \textbf{Prompt 1:} The prompt specifies that only the initial letter of the word ``Test'' should be capitalized, but the response is entirely in uppercase. Here, {\ours} assigns a negative reward to capture the mismatch, whereas DPO incorrectly assigns a positive reward.
    \item \textbf{Prompt 2:} The correct response is \textit{one dog}, yet DPO erroneously assigns a positive reward to the incorrect answer. In contrast, {\ours} accurately reflects the error by assigning a negative reward.
    \item \textbf{Prompt 3:} The correct answer is 4, and the response matches. While DPO fails to assign a positive reward to this response, {\ours} correctly assigns a positive reward.
\end{itemize}
These examples underscore the effectiveness of {\ours} in achieving precise token-level reward assignments, demonstrating its superiority in addressing discrepancies that DPO overlooks.

\definecolor{softred}{RGB}{240, 90, 84}
\definecolor{purpleblue}{RGB}{114, 90, 193}
\definecolor{lightpurple}{RGB}{182, 159, 228}

\begin{figure}
    \centering
    \includegraphics[width=0.95\linewidth]{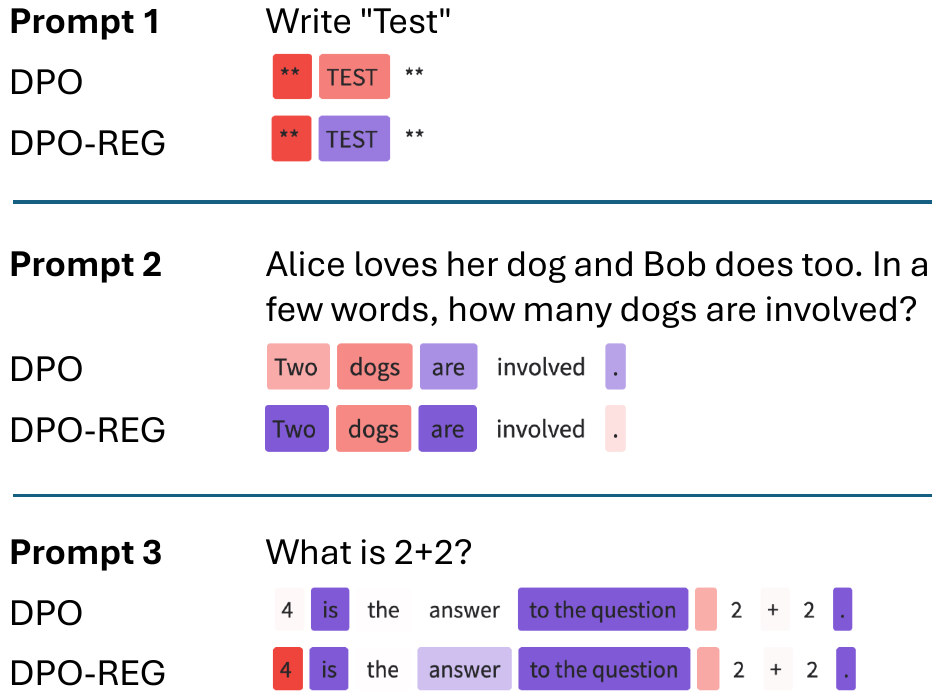}
    \caption{Case study on the token-level rewards learned by DPO and DPO-REG, where \colorbox{softred}{red} means positive reward and \colorbox{lightpurple}{blue} means negative reward. We use Llama-3-Instruct (8B) as the base model.}
    \label{fig:case_study}
\end{figure}

\section{Conclusion}

In this study, we addressed the challenge of token-level credit assignment in preference optimization by introducing {\ours}, a novel method that leverages self-generated token-level rewards derived through opposite prompting as regularization. This approach allows for more effective and fine-grained credit assignment, seamlessly integrating with sequence-level preference optimization to enhance alignment with human preferences.
Experiments on instruction-following benchmarks including Alpaca Eval 2 and Arena-Hard, as well as our qualitative case study, demonstrate that {\ours} not only enhances alignment performance but also achieves more precise token-level credit assignment.

\section*{Limitations}
\smallskip
\noindent\textbf{Lack of quantitative results on token-level credit assignment.}
In this paper, we only provide qualitative results on the token-level rewards learned by {\ours}.
Therefore, there lacks a systematic and more rigorous study on the accuracy of token-level rewards.
While we believe this analysis will provide valuable insights, to the best of our knowledge, no benchmarks currently exist for evaluating token-level rewards. Future work should focus on constructing dedicated evaluation datasets to facilitate quantitative assessment of token-level reward accuracy.

\smallskip
\noindent\textbf{Rewards in other levels.}
Our approach focuses on utilizing rewards at the token and sequence levels. However, intermediate levels, such as step-level and span-level rewards, also provide useful information for alignment tasks and have been widely applied, especially in math and coding problems.
The current method does not account for these intermediate reward levels. Future research could explore methods that incorporate multiple levels of rewards, potentially enhancing the flexibility and effectiveness of preference optimization.

\bibliography{acl_latex}

\begin{thebibliography}{60}
\providecommand{\natexlab}[1]{#1}

\bibitem[{Azar et~al.(2024)Azar, Guo, Piot, Munos, Rowland, Valko, and Calandriello}]{azar2024general}
Mohammad~Gheshlaghi Azar, Zhaohan~Daniel Guo, Bilal Piot, Remi Munos, Mark Rowland, Michal Valko, and Daniele Calandriello. 2024.
\newblock A general theoretical paradigm to understand learning from human preferences.
\newblock In \emph{International Conference on Artificial Intelligence and Statistics}, pages 4447--4455. PMLR.

\bibitem[{Bai et~al.(2022)Bai, Jones, Ndousse, Askell, Chen, DasSarma, Drain, Fort, Ganguli, Henighan et~al.}]{bai2022training}
Yuntao Bai, Andy Jones, Kamal Ndousse, Amanda Askell, Anna Chen, Nova DasSarma, Dawn Drain, Stanislav Fort, Deep Ganguli, Tom Henighan, et~al. 2022.
\newblock Training a helpful and harmless assistant with reinforcement learning from human feedback.
\newblock \emph{arXiv preprint arXiv:2204.05862}.

\bibitem[{Bradley and Terry(1952)}]{bradley1952rank}
Ralph~Allan Bradley and Milton~E Terry. 1952.
\newblock Rank analysis of incomplete block designs: I. the method of paired comparisons.
\newblock \emph{Biometrika}, 39(3/4):324--345.

\bibitem[{Chan et~al.(2024)Chan, Sun, Holt, and van~der Schaar}]{chan2024dense}
Alex~J Chan, Hao Sun, Samuel Holt, and Mihaela van~der Schaar. 2024.
\newblock Dense reward for free in reinforcement learning from human feedback.
\newblock \emph{arXiv preprint arXiv:2402.00782}.

\bibitem[{Chen et~al.(2018)Chen, Badrinarayanan, Lee, and Rabinovich}]{chen2018gradnorm}
Zhao Chen, Vijay Badrinarayanan, Chen-Yu Lee, and Andrew Rabinovich. 2018.
\newblock Gradnorm: Gradient normalization for adaptive loss balancing in deep multitask networks.
\newblock In \emph{International conference on machine learning}, pages 794--803. PMLR.

\bibitem[{Chen et~al.(2024)Chen, Zhou, Zhao, Wan, Zhang, Zhang, and Wen}]{chen2024improving}
Zhipeng Chen, Kun Zhou, Wayne~Xin Zhao, Junchen Wan, Fuzheng Zhang, Di~Zhang, and Ji-Rong Wen. 2024.
\newblock Improving large language models via fine-grained reinforcement learning with minimum editing constraint.
\newblock \emph{arXiv preprint arXiv:2401.06081}.

\bibitem[{Chiang et~al.(2024)Chiang, Zheng, Sheng, Angelopoulos, Li, Li, Zhang, Zhu, Jordan, Gonzalez et~al.}]{chiang2024chatbot}
Wei-Lin Chiang, Lianmin Zheng, Ying Sheng, Anastasios~Nikolas Angelopoulos, Tianle Li, Dacheng Li, Hao Zhang, Banghua Zhu, Michael Jordan, Joseph~E Gonzalez, et~al. 2024.
\newblock Chatbot arena: An open platform for evaluating llms by human preference.
\newblock \emph{arXiv preprint arXiv:2403.04132}.

\bibitem[{Chowdhury et~al.(2024)Chowdhury, Kini, and Natarajan}]{chowdhury2024provably}
Sayak~Ray Chowdhury, Anush Kini, and Nagarajan Natarajan. 2024.
\newblock Provably robust dpo: Aligning language models with noisy feedback.
\newblock \emph{arXiv preprint arXiv:2403.00409}.

\bibitem[{Christiano et~al.(2017)Christiano, Leike, Brown, Martic, Legg, and Amodei}]{christiano2017deep}
Paul~F Christiano, Jan Leike, Tom Brown, Miljan Martic, Shane Legg, and Dario Amodei. 2017.
\newblock Deep reinforcement learning from human preferences.
\newblock \emph{Advances in neural information processing systems}, 30.

\bibitem[{Chung et~al.(2024)Chung, Hou, Longpre, Zoph, Tay, Fedus, Li, Wang, Dehghani, Brahma et~al.}]{chung2024scaling}
Hyung~Won Chung, Le~Hou, Shayne Longpre, Barret Zoph, Yi~Tay, William Fedus, Yunxuan Li, Xuezhi Wang, Mostafa Dehghani, Siddhartha Brahma, et~al. 2024.
\newblock Scaling instruction-finetuned language models.
\newblock \emph{Journal of Machine Learning Research}, 25(70):1--53.

\bibitem[{Cui et~al.(2023)Cui, Yuan, Ding, Yao, Zhu, Ni, Xie, Liu, and Sun}]{cui2023ultrafeedback}
Ganqu Cui, Lifan Yuan, Ning Ding, Guanming Yao, Wei Zhu, Yuan Ni, Guotong Xie, Zhiyuan Liu, and Maosong Sun. 2023.
\newblock Ultrafeedback: Boosting language models with high-quality feedback.
\newblock \emph{arXiv preprint arXiv:2310.01377}.

\bibitem[{Dong et~al.(2023)Dong, Xiong, Goyal, Pan, Diao, Zhang, Shum, and Zhang}]{dong2023raft}
Hanze Dong, Wei Xiong, Deepanshu Goyal, Rui Pan, Shizhe Diao, Jipeng Zhang, Kashun Shum, and Tong Zhang. 2023.
\newblock Raft: Reward ranked finetuning for generative foundation model alignment.
\newblock \emph{arXiv preprint arXiv:2304.06767}.

\bibitem[{Dubey et~al.(2024)Dubey, Jauhri, Pandey, Kadian, Al-Dahle, Letman, Mathur, Schelten, Yang, Fan et~al.}]{dubey2024llama}
Abhimanyu Dubey, Abhinav Jauhri, Abhinav Pandey, Abhishek Kadian, Ahmad Al-Dahle, Aiesha Letman, Akhil Mathur, Alan Schelten, Amy Yang, Angela Fan, et~al. 2024.
\newblock The llama 3 herd of models.
\newblock \emph{arXiv preprint arXiv:2407.21783}.

\bibitem[{Dubois et~al.(2024)Dubois, Galambosi, Liang, and Hashimoto}]{dubois2024length}
Yann Dubois, Bal{\'a}zs Galambosi, Percy Liang, and Tatsunori~B Hashimoto. 2024.
\newblock Length-controlled alpacaeval: A simple way to debias automatic evaluators.
\newblock \emph{arXiv preprint arXiv:2404.04475}.

\bibitem[{Engstrom et~al.(2019)Engstrom, Ilyas, Santurkar, Tsipras, Janoos, Rudolph, and Madry}]{engstrom2019implementation}
Logan Engstrom, Andrew Ilyas, Shibani Santurkar, Dimitris Tsipras, Firdaus Janoos, Larry Rudolph, and Aleksander Madry. 2019.
\newblock Implementation matters in deep rl: A case study on ppo and trpo.
\newblock In \emph{International conference on learning representations}.

\bibitem[{Ethayarajh et~al.(2024)Ethayarajh, Xu, Muennighoff, Jurafsky, and Kiela}]{ethayarajh2024kto}
Kawin Ethayarajh, Winnie Xu, Niklas Muennighoff, Dan Jurafsky, and Douwe Kiela. 2024.
\newblock Kto: Model alignment as prospect theoretic optimization.
\newblock \emph{arXiv preprint arXiv:2402.01306}.

\bibitem[{Gao et~al.(2023)Gao, Schulman, and Hilton}]{gao2023scaling}
Leo Gao, John Schulman, and Jacob Hilton. 2023.
\newblock Scaling laws for reward model overoptimization.
\newblock In \emph{International Conference on Machine Learning}, pages 10835--10866. PMLR.

\bibitem[{Glaese et~al.(2022)Glaese, McAleese, Tr{\k{e}}bacz, Aslanides, Firoiu, Ewalds, Rauh, Weidinger, Chadwick, Thacker et~al.}]{glaese2022improving}
Amelia Glaese, Nat McAleese, Maja Tr{\k{e}}bacz, John Aslanides, Vlad Firoiu, Timo Ewalds, Maribeth Rauh, Laura Weidinger, Martin Chadwick, Phoebe Thacker, et~al. 2022.
\newblock Improving alignment of dialogue agents via targeted human judgements.
\newblock \emph{arXiv preprint arXiv:2209.14375}.

\bibitem[{Guo et~al.(2023)Guo, Zhao, Tang, Zhao, and Wen}]{guo2023beyond}
Geyang Guo, Ranchi Zhao, Tianyi Tang, Wayne~Xin Zhao, and Ji-Rong Wen. 2023.
\newblock Beyond imitation: Leveraging fine-grained quality signals for alignment.
\newblock \emph{arXiv preprint arXiv:2311.04072}.

\bibitem[{Guo et~al.(2024)Guo, Zhang, Liu, Liu, Khalman, Llinares, Rame, Mesnard, Zhao, Piot et~al.}]{guo2024direct}
Shangmin Guo, Biao Zhang, Tianlin Liu, Tianqi Liu, Misha Khalman, Felipe Llinares, Alexandre Rame, Thomas Mesnard, Yao Zhao, Bilal Piot, et~al. 2024.
\newblock Direct language model alignment from online ai feedback.
\newblock \emph{arXiv preprint arXiv:2402.04792}.

\bibitem[{Huang et~al.(2024)Huang, Zhou, Wang, Morstatter, Zhang, Poon, and Chen}]{huang2024offset}
James~Y Huang, Wenxuan Zhou, Fei Wang, Fred Morstatter, Sheng Zhang, Hoifung Poon, and Muhao Chen. 2024.
\newblock Offset unlearning for large language models.
\newblock \emph{arXiv preprint arXiv:2404.11045}.

\bibitem[{Kasai et~al.(2022)Kasai, Sakaguchi, Bras, Peng, Lu, Radev, Choi, and Smith}]{kasai2022twist}
Jungo Kasai, Keisuke Sakaguchi, Ronan~Le Bras, Hao Peng, Ximing Lu, Dragomir Radev, Yejin Choi, and Noah~A Smith. 2022.
\newblock Twist decoding: Diverse generators guide each other.
\newblock \emph{arXiv preprint arXiv:2205.09273}.

\bibitem[{Kim et~al.(2024)Kim, Kim, Lee, and Yun}]{kim2024instructive}
Taehyeon Kim, Joonkee Kim, Gihun Lee, and Se-Young Yun. 2024.
\newblock Instructive decoding: Instruction-tuned large language models are self-refiner from noisy instructions.
\newblock In \emph{The Twelfth International Conference on Learning Representations}.

\bibitem[{Lambert et~al.(2024)Lambert, Pyatkin, Morrison, Miranda, Lin, Chandu, Dziri, Kumar, Zick, Choi et~al.}]{lambert2024rewardbench}
Nathan Lambert, Valentina Pyatkin, Jacob Morrison, LJ~Miranda, Bill~Yuchen Lin, Khyathi Chandu, Nouha Dziri, Sachin Kumar, Tom Zick, Yejin Choi, et~al. 2024.
\newblock Rewardbench: Evaluating reward models for language modeling.
\newblock \emph{arXiv preprint arXiv:2403.13787}.

\bibitem[{Lee et~al.(2023)Lee, Phatale, Mansoor, Mesnard, Ferret, Lu, Bishop, Hall, Carbune, Rastogi et~al.}]{lee2023rlaif}
Harrison Lee, Samrat Phatale, Hassan Mansoor, Thomas Mesnard, Johan Ferret, Kellie Lu, Colton Bishop, Ethan Hall, Victor Carbune, Abhinav Rastogi, et~al. 2023.
\newblock Rlaif: Scaling reinforcement learning from human feedback with ai feedback.
\newblock \emph{arXiv preprint arXiv:2309.00267}.

\bibitem[{Li et~al.(2024)Li, Chiang, Frick, Dunlap, Wu, Zhu, Gonzalez, and Stoica}]{li2024crowdsourced}
Tianle Li, Wei-Lin Chiang, Evan Frick, Lisa Dunlap, Tianhao Wu, Banghua Zhu, Joseph~E Gonzalez, and Ion Stoica. 2024.
\newblock From crowdsourced data to high-quality benchmarks: Arena-hard and benchbuilder pipeline.
\newblock \emph{arXiv preprint arXiv:2406.11939}.

\bibitem[{Li et~al.(2022)Li, Holtzman, Fried, Liang, Eisner, Hashimoto, Zettlemoyer, and Lewis}]{li2022contrastive}
Xiang~Lisa Li, Ari Holtzman, Daniel Fried, Percy Liang, Jason Eisner, Tatsunori Hashimoto, Luke Zettlemoyer, and Mike Lewis. 2022.
\newblock Contrastive decoding: Open-ended text generation as optimization.
\newblock \emph{arXiv preprint arXiv:2210.15097}.

\bibitem[{Li et~al.(2023)Li, Zhang, Dubois, Taori, Gulrajani, Guestrin, Liang, and Hashimoto}]{alpaca_eval}
Xuechen Li, Tianyi Zhang, Yann Dubois, Rohan Taori, Ishaan Gulrajani, Carlos Guestrin, Percy Liang, and Tatsunori~B. Hashimoto. 2023.
\newblock Alpacaeval: An automatic evaluator of instruction-following models.
\newblock \url{https://github.com/tatsu-lab/alpaca_eval}.

\bibitem[{Lightman et~al.(2023)Lightman, Kosaraju, Burda, Edwards, Baker, Lee, Leike, Schulman, Sutskever, and Cobbe}]{lightman2023let}
Hunter Lightman, Vineet Kosaraju, Yura Burda, Harri Edwards, Bowen Baker, Teddy Lee, Jan Leike, John Schulman, Ilya Sutskever, and Karl Cobbe. 2023.
\newblock Let's verify step by step.
\newblock \emph{arXiv preprint arXiv:2305.20050}.

\bibitem[{Liu et~al.(2023)Liu, Iter, Xu, Wang, Xu, and Zhu}]{liu-etal-2023-g}
Yang Liu, Dan Iter, Yichong Xu, Shuohang Wang, Ruochen Xu, and Chenguang Zhu. 2023.
\newblock \href {https://doi.org/10.18653/v1/2023.emnlp-main.153} {{G}-eval: {NLG} evaluation using gpt-4 with better human alignment}.
\newblock In \emph{Proceedings of the 2023 Conference on Empirical Methods in Natural Language Processing}, pages 2511--2522, Singapore. Association for Computational Linguistics.

\bibitem[{Lu et~al.(2022)Lu, Welleck, Hessel, Jiang, Qin, West, Ammanabrolu, and Choi}]{lu2022quark}
Ximing Lu, Sean Welleck, Jack Hessel, Liwei Jiang, Lianhui Qin, Peter West, Prithviraj Ammanabrolu, and Yejin Choi. 2022.
\newblock Quark: Controllable text generation with reinforced unlearning.
\newblock \emph{Advances in neural information processing systems}, 35:27591--27609.

\bibitem[{Madaan et~al.(2024)Madaan, Tandon, Gupta, Hallinan, Gao, Wiegreffe, Alon, Dziri, Prabhumoye, Yang et~al.}]{madaan2024self}
Aman Madaan, Niket Tandon, Prakhar Gupta, Skyler Hallinan, Luyu Gao, Sarah Wiegreffe, Uri Alon, Nouha Dziri, Shrimai Prabhumoye, Yiming Yang, et~al. 2024.
\newblock Self-refine: Iterative refinement with self-feedback.
\newblock \emph{Advances in Neural Information Processing Systems}, 36.

\bibitem[{Meng et~al.(2024)Meng, Xia, and Chen}]{meng2024simpo}
Yu~Meng, Mengzhou Xia, and Danqi Chen. 2024.
\newblock Simpo: Simple preference optimization with a reference-free reward.
\newblock \emph{arXiv preprint arXiv:2405.14734}.

\bibitem[{Ouyang et~al.(2022)Ouyang, Wu, Jiang, Almeida, Wainwright, Mishkin, Zhang, Agarwal, Slama, Ray et~al.}]{ouyang2022training}
Long Ouyang, Jeffrey Wu, Xu~Jiang, Diogo Almeida, Carroll Wainwright, Pamela Mishkin, Chong Zhang, Sandhini Agarwal, Katarina Slama, Alex Ray, et~al. 2022.
\newblock Training language models to follow instructions with human feedback.
\newblock \emph{Advances in neural information processing systems}, 35:27730--27744.

\bibitem[{Rafailov et~al.(2024)Rafailov, Hejna, Park, and Finn}]{rafailov2024r}
Rafael Rafailov, Joey Hejna, Ryan Park, and Chelsea Finn. 2024.
\newblock From $ r $ to $ q^* $: Your language model is secretly a q-function.
\newblock \emph{arXiv preprint arXiv:2404.12358}.

\bibitem[{Rafailov et~al.(2023)Rafailov, Sharma, Mitchell, Manning, Ermon, and Finn}]{rafailov2024direct}
Rafael Rafailov, Archit Sharma, Eric Mitchell, Christopher~D Manning, Stefano Ermon, and Chelsea Finn. 2023.
\newblock Direct preference optimization: Your language model is secretly a reward model.
\newblock \emph{Advances in Neural Information Processing Systems}, 36.

\bibitem[{Schulman et~al.(2017)Schulman, Wolski, Dhariwal, Radford, and Klimov}]{schulman2017proximal}
John Schulman, Filip Wolski, Prafulla Dhariwal, Alec Radford, and Oleg Klimov. 2017.
\newblock Proximal policy optimization algorithms.
\newblock \emph{arXiv preprint arXiv:1707.06347}.

\bibitem[{Team et~al.(2024)Team, Mesnard, Hardin, Dadashi, Bhupatiraju, Pathak, Sifre, Rivi{\`e}re, Kale, Love et~al.}]{team2024gemma}
Gemma Team, Thomas Mesnard, Cassidy Hardin, Robert Dadashi, Surya Bhupatiraju, Shreya Pathak, Laurent Sifre, Morgane Rivi{\`e}re, Mihir~Sanjay Kale, Juliette Love, et~al. 2024.
\newblock Gemma: Open models based on gemini research and technology.
\newblock \emph{arXiv preprint arXiv:2403.08295}.

\bibitem[{Tunstall et~al.(2023)Tunstall, Beeching, Lambert, Rajani, Rasul, Belkada, Huang, von Werra, Fourrier, Habib et~al.}]{tunstall2023zephyr}
Lewis Tunstall, Edward Beeching, Nathan Lambert, Nazneen Rajani, Kashif Rasul, Younes Belkada, Shengyi Huang, Leandro von Werra, Cl{\'e}mentine Fourrier, Nathan Habib, et~al. 2023.
\newblock Zephyr: Direct distillation of lm alignment.
\newblock \emph{arXiv preprint arXiv:2310.16944}.

\bibitem[{Uesato et~al.(2022)Uesato, Kushman, Kumar, Song, Siegel, Wang, Creswell, Irving, and Higgins}]{uesato2022solving}
Jonathan Uesato, Nate Kushman, Ramana Kumar, Francis Song, Noah Siegel, Lisa Wang, Antonia Creswell, Geoffrey Irving, and Irina Higgins. 2022.
\newblock Solving math word problems with process-and outcome-based feedback.
\newblock \emph{arXiv preprint arXiv:2211.14275}.

\bibitem[{Wang et~al.(2024{\natexlab{a}})Wang, Zheng, Chen, Liu, Dou, Huang, Shen, Jin, Zhou, Shi et~al.}]{wang2024secrets}
Binghai Wang, Rui Zheng, Lu~Chen, Yan Liu, Shihan Dou, Caishuang Huang, Wei Shen, Senjie Jin, Enyu Zhou, Chenyu Shi, et~al. 2024{\natexlab{a}}.
\newblock Secrets of rlhf in large language models part ii: Reward modeling.
\newblock \emph{arXiv preprint arXiv:2401.06080}.

\bibitem[{Wang et~al.(2024{\natexlab{b}})Wang, Zhou, Huang, Xu, Zhang, Poon, and Chen}]{wang2024mdpo}
Fei Wang, Wenxuan Zhou, James~Y Huang, Nan Xu, Sheng Zhang, Hoifung Poon, and Muhao Chen. 2024{\natexlab{b}}.
\newblock mdpo: Conditional preference optimization for multimodal large language models.
\newblock \emph{arXiv preprint arXiv:2406.11839}.

\bibitem[{Wang et~al.(2024{\natexlab{c}})Wang, Lin, Xiong, Yang, Diao, Qiu, Zhao, and Zhang}]{wang2024arithmetic}
Haoxiang Wang, Yong Lin, Wei Xiong, Rui Yang, Shizhe Diao, Shuang Qiu, Han Zhao, and Tong Zhang. 2024{\natexlab{c}}.
\newblock Arithmetic control of llms for diverse user preferences: Directional preference alignment with multi-objective rewards.
\newblock In \emph{ACL}.

\bibitem[{Wang et~al.(2024{\natexlab{d}})Wang, Xiong, Xie, Zhao, and Zhang}]{ArmoRM}
Haoxiang Wang, Wei Xiong, Tengyang Xie, Han Zhao, and Tong Zhang. 2024{\natexlab{d}}.
\newblock Interpretable preferences via multi-objective reward modeling and mixture-of-experts.
\newblock \emph{arXiv preprint arXiv:2406.12845}.

\bibitem[{Wang et~al.(2024{\natexlab{e}})Wang, Kulikov, Golovneva, Yu, Yuan, Dwivedi-Yu, Pang, Fazel-Zarandi, Weston, and Li}]{wang2024self}
Tianlu Wang, Ilia Kulikov, Olga Golovneva, Ping Yu, Weizhe Yuan, Jane Dwivedi-Yu, Richard~Yuanzhe Pang, Maryam Fazel-Zarandi, Jason Weston, and Xian Li. 2024{\natexlab{e}}.
\newblock Self-taught evaluators.
\newblock \emph{arXiv preprint arXiv:2408.02666}.

\bibitem[{Wu et~al.(2024)Wu, Hu, Shi, Dziri, Suhr, Ammanabrolu, Smith, Ostendorf, and Hajishirzi}]{wu2024fine}
Zeqiu Wu, Yushi Hu, Weijia Shi, Nouha Dziri, Alane Suhr, Prithviraj Ammanabrolu, Noah~A Smith, Mari Ostendorf, and Hannaneh Hajishirzi. 2024.
\newblock Fine-grained human feedback gives better rewards for language model training.
\newblock \emph{Advances in Neural Information Processing Systems}, 36.

\bibitem[{Xiong et~al.(2024)Xiong, Dong, Ye, Wang, Zhong, Ji, Jiang, and Zhang}]{xiong2024iterative}
Wei Xiong, Hanze Dong, Chenlu Ye, Ziqi Wang, Han Zhong, Heng Ji, Nan Jiang, and Tong Zhang. 2024.
\newblock \href {https://arxiv.org/abs/2312.11456} {Iterative preference learning from human feedback: Bridging theory and practice for rlhf under kl-constraint}.
\newblock \emph{Preprint}, arXiv:2312.11456.

\bibitem[{Xu et~al.(2024)Xu, Sharaf, Chen, Tan, Shen, Van~Durme, Murray, and Kim}]{xucontrastive}
Haoran Xu, Amr Sharaf, Yunmo Chen, Weiting Tan, Lingfeng Shen, Benjamin Van~Durme, Kenton Murray, and Young~Jin Kim. 2024.
\newblock Contrastive preference optimization: Pushing the boundaries of llm performance in machine translation.
\newblock In \emph{Forty-first International Conference on Machine Learning}.

\bibitem[{Yang et~al.(2024{\natexlab{a}})Yang, Liu, Xie, Huang, Min, and Ananiadou}]{yang2024selective}
Kailai Yang, Zhiwei Liu, Qianqian Xie, Jimin Huang, Erxue Min, and Sophia Ananiadou. 2024{\natexlab{a}}.
\newblock Selective preference optimization via token-level reward function estimation.
\newblock \emph{arXiv preprint arXiv:2408.13518}.

\bibitem[{Yang et~al.(2024{\natexlab{b}})Yang, Zhang, Xia, Feng, Xiong, and Zhou}]{yang2024preference}
Shentao Yang, Shujian Zhang, Congying Xia, Yihao Feng, Caiming Xiong, and Mingyuan Zhou. 2024{\natexlab{b}}.
\newblock Preference-grounded token-level guidance for language model fine-tuning.
\newblock \emph{Advances in Neural Information Processing Systems}, 36.

\bibitem[{Yoon et~al.(2024)Yoon, Yoon, Eom, Han, Nam, Jo, On, Hasegawa-Johnson, Kim, and Yoo}]{yoon2024tlcr}
Eunseop Yoon, Hee~Suk Yoon, SooHwan Eom, Gunsoo Han, Daniel~Wontae Nam, Daejin Jo, Kyoung-Woon On, Mark~A Hasegawa-Johnson, Sungwoong Kim, and Chang~D Yoo. 2024.
\newblock Tlcr: Token-level continuous reward for fine-grained reinforcement learning from human feedback.
\newblock \emph{arXiv preprint arXiv:2407.16574}.

\bibitem[{Zeng et~al.(2024)Zeng, Liu, Ma, Yang, Zhang, and Wang}]{zeng2024token}
Yongcheng Zeng, Guoqing Liu, Weiyu Ma, Ning Yang, Haifeng Zhang, and Jun Wang. 2024.
\newblock Token-level direct preference optimization.
\newblock \emph{arXiv preprint arXiv:2404.11999}.

\bibitem[{Zhao et~al.(2024{\natexlab{a}})Zhao, Yang, Pang, Du, Li, Wang, and Wang}]{zhao2024weak}
Xuandong Zhao, Xianjun Yang, Tianyu Pang, Chao Du, Lei Li, Yu-Xiang Wang, and William~Yang Wang. 2024{\natexlab{a}}.
\newblock Weak-to-strong jailbreaking on large language models.
\newblock \emph{arXiv preprint arXiv:2401.17256}.

\bibitem[{Zhao et~al.(2023)Zhao, Joshi, Liu, Khalman, Saleh, and Liu}]{zhao2023slic}
Yao Zhao, Rishabh Joshi, Tianqi Liu, Misha Khalman, Mohammad Saleh, and Peter~J Liu. 2023.
\newblock Slic-hf: Sequence likelihood calibration with human feedback.
\newblock \emph{arXiv preprint arXiv:2305.10425}.

\bibitem[{Zhao et~al.(2024{\natexlab{b}})Zhao, Zhang, Xu, Hu, Zhang, Du, Guo, and Chen}]{zhao2024adversarial}
Zhengyue Zhao, Xiaoyun Zhang, Kaidi Xu, Xing Hu, Rui Zhang, Zidong Du, Qi~Guo, and Yunji Chen. 2024{\natexlab{b}}.
\newblock Adversarial contrastive decoding: Boosting safety alignment of large language models via opposite prompt optimization.
\newblock \emph{arXiv preprint arXiv:2406.16743}.

\bibitem[{Zheng et~al.(2024)Zheng, Wang, Ji, Huang, and Peng}]{zheng2024weak}
Chujie Zheng, Ziqi Wang, Heng Ji, Minlie Huang, and Nanyun Peng. 2024.
\newblock Weak-to-strong extrapolation expedites alignment.
\newblock \emph{arXiv preprint arXiv:2404.16792}.

\bibitem[{Zhong et~al.(2024)Zhong, Feng, Xiong, Zhao, He, Bian, and Wang}]{zhong2024dpo}
Han Zhong, Guhao Feng, Wei Xiong, Li~Zhao, Di~He, Jiang Bian, and Liwei Wang. 2024.
\newblock Dpo meets ppo: Reinforced token optimization for rlhf.
\newblock \emph{arXiv preprint arXiv:2404.18922}.

\bibitem[{Zhou et~al.(2024{\natexlab{a}})Zhou, Agrawal, Zhang, Indurthi, Zhao, Song, Xu, and Zhu}]{zhou2024wpo}
Wenxuan Zhou, Ravi Agrawal, Shujian Zhang, Sathish~Reddy Indurthi, Sanqiang Zhao, Kaiqiang Song, Silei Xu, and Chenguang Zhu. 2024{\natexlab{a}}.
\newblock Wpo: Enhancing rlhf with weighted preference optimization.
\newblock \emph{arXiv preprint arXiv:2406.11827}.

\bibitem[{Zhou et~al.(2024{\natexlab{b}})Zhou, Liu, Liu, Dong, Yang, and Qiao}]{zhou2024weak}
Zhanhui Zhou, Zhixuan Liu, Jie Liu, Zhichen Dong, Chao Yang, and Yu~Qiao. 2024{\natexlab{b}}.
\newblock Weak-to-strong search: Align large language models via searching over small language models.
\newblock \emph{arXiv preprint arXiv:2405.19262}.

\bibitem[{Ziegler et~al.(2019)Ziegler, Stiennon, Wu, Brown, Radford, Amodei, Christiano, and Irving}]{ziegler2019fine}
Daniel~M Ziegler, Nisan Stiennon, Jeffrey Wu, Tom~B Brown, Alec Radford, Dario Amodei, Paul Christiano, and Geoffrey Irving. 2019.
\newblock Fine-tuning language models from human preferences.
\newblock \emph{arXiv preprint arXiv:1909.08593}.

\end{thebibliography}

\end{document}